\documentclass[10pt,twocolumn,a4paper]{article}
\usepackage[T1]{fontenc}
\usepackage[utf8]{inputenc}
\usepackage{newtxtext,newtxmath}
\usepackage[a4paper,top=18mm,bottom=20mm,left=18mm,right=18mm,columnsep=6mm]{geometry}
\usepackage{graphicx}
\usepackage{amsmath}
\usepackage{microtype}
\usepackage[numbers,sort&compress]{natbib}
\usepackage[font=small,labelfont=bf,labelsep=period]{caption}
\usepackage[hidelinks]{hyperref}
\hypersetup{pdftitle={ZoomSEM: accelerating large-scale brain connectomics with intelligent SEM super-resolution},
 pdfauthor={Jiateng Shou, Zhiwei Xiong, Ruikang Xu, Zhicheng Wang, Peiyao Shi, Te Shi, Haiqun Jin, Chunying Yin, Jun Guo, Manfred Auer, Hanchuan Peng, Ruobing Zhang, Feng Wu}}
\begin{document}
\twocolumn[{
\begin{center}
{\LARGE\bfseries ZoomSEM: accelerating large-scale brain connectomics\\[3pt]
with intelligent SEM super-resolution\par}
\vspace{10pt}
{\normalsize
Jiateng Shou\textsuperscript{1,2}, Zhiwei Xiong\textsuperscript{1,2,*}, Ruikang Xu\textsuperscript{1,2}, Zhicheng Wang\textsuperscript{1,2}, Peiyao Shi\textsuperscript{3},\\[3pt]
Te Shi\textsuperscript{2}, Haiqun Jin\textsuperscript{2}, Chunying Yin\textsuperscript{2}, Jun Guo\textsuperscript{2}, Manfred Auer\textsuperscript{4},\\[3pt]
Hanchuan Peng\textsuperscript{3}, Ruobing Zhang\textsuperscript{3,2,*}, Feng Wu\textsuperscript{1,2,*}\par}
\vspace{7pt}
{\footnotesize
\textsuperscript{1}MoE Key Laboratory of Brain-inspired Intelligent Perception and Cognition, University of Science and Technology of China, Hefei, China.\par
\textsuperscript{2}Anhui Province Key Laboratory of Biomedical Imaging and Intelligent Processing, Institute of Artificial Intelligence, Hefei Comprehensive National Science Center, Hefei, China.\par
\textsuperscript{3}Institute for Brain and Intelligence, Fudan University, Shanghai, China.\par
\textsuperscript{4}Department of Biomedical Engineering, School of Biological Sciences and Medical Engineering, Southeast University, Nanjing, China.\par}
\vspace{5pt}
{\footnotesize\textsuperscript{*}Correspondence: \href{mailto:zwxiong@ustc.edu.cn}{zwxiong@ustc.edu.cn}, \href{mailto:rbzhang@fudan.edu.cn}{rbzhang@fudan.edu.cn}, \href{mailto:fengwu@ustc.edu.cn}{fengwu@ustc.edu.cn}\par}
\end{center}
\vspace{3pt}
\begin{minipage}{\textwidth}
\small
\noindent\textbf{Abstract.} Serial electron microscopy (EM) is essential for large-volume biological imaging and connectomics, yet its throughput remains a major bottleneck, with cubic-millimeter-scale brain imaging requiring months or years. Existing deep learning--based EM super-resolution methods are limited to small upscaling factors or rely on synthetic data, offering modest acceleration and limited practical impact. We introduce ZoomSEM, a deep learning-based super-resolution method to dramatically accelerate SEM imaging. Trained on real low- and high-resolution scanning electron microscopy (SEM) pairs, ZoomSEM enables reliable recovery of 4\ensuremath{\times}, 8\ensuremath{\times}, and up to 16\ensuremath{\times} higher-resolution images from fast low-resolution scans while preserving fine neuronal structures critical for connectomic analysis, achieving up to 30-fold effective acceleration. Extensive experiments across datasets of diverse biological samples demonstrate robust generalization under realistic low-resolution acquisition. This advance enables high-resolution, large-volume serial EM imaging at unprecedented speed, poised to transform connectomics and related fields.

\vspace{6pt}
\end{minipage}
\vspace{14pt}
}]

\section*{Introduction}
High-resolution serial electron microscopy (EM) has become a pivotal tool in modern biological research, particularly for detailed mapping of brain connectomes at synaptic resolution \cite{peddie2022volume,xu2021open,iudin2016empiar}. Scanning electron microscopes (SEMs) routinely achieve nanometer-level spatial resolution, revealing ultrastructural details far beyond optical imaging. However, imaging large volumes at this scale is painfully slow. Standard single-beam SEMs scan sequentially one pixel per tens or hundreds of nanoseconds. State-of-the-art multi-beam SEMs (e.g. 61 or 91 beams) can theoretically capture $\sim$1.2 gigapixel per second, but mechanical overhead (stage movement, auto-focusing, etc.) limits effective throughput to only $\sim$0.2 Gpix/s \cite{eberle2015high,zheng2024fast}. Consequently, imaging a cubic millimeter of brain tissue at nanometer resolution would typically take many months or even years on current instruments~\cite{shapson2024petavoxel}. Reducing pixel dwell time increases throughput at the expense of signal-to-noise ratio (SNR), i.e., a speed-SNR trade-off. Such constraints pose a substantial bottleneck to the application of serial EM, especially in large-scale brain connectomics.

To alleviate this, computational EM image resolution enhancement, often referred to as EM super-resolution, has emerged as a promising strategy. Recent studies have shown that deep neural networks \cite{lecun2015deep} trained on paired low/high resolution EM images can infer unresolved features in coarse scans \cite{de2019resolution,reid2022multi,xie2022augmented,qian2020effective,fang2021deep,shou2024learning,lu2024diffusion}. Despite notable progress in deep learning-based approaches, several limitations continue to hinder their application in practical EM imaging. First, most existing methods \cite{de2019resolution,xie2022augmented,qian2020effective,fang2021deep,lu2024diffusion} are limited to small upscaling factors and degrade significantly in performance at higher magnifications.
Although a few studies have explored large-factor super-resolution \cite{fang2021deep,shou2024learning}, they typically rely on synthetic training data generated by artificially degrading high-resolution EM images, an approach that poorly reflects real EM degradation processes.
Second, current methods often fail to fully exploit spatial continuity across serial EM sections. Many adopt multi-frame techniques \cite{ranjan2017optical,teed2020raft,dai2017deformable,zhu2019deformable} from video super-resolution \cite{wang2019edvr,chan2021basicvsr,chan2022basicvsr++,xu2024enhancing} that rely on optical flow but overlook the structural scale variance characteristic of serial EM images.
Third, previous methods primarily focused on algorithmic design and improving performance on small-scale datasets, without considering practical applicability to large-scale data.

In this work, we present ZoomSEM, a novel deep-learning network that significantly enhances SEM imaging efficiency by computationally restoring high-resolution ultrastructures from low-resolution SEM images across multiple scales. ZoomSEM is designed to support 4$\times$, 8$\times$, and 16$\times$ enhancements in lateral resolution, allowing correspondingly coarser scans (i.e., 4$\times$, 8$\times$, and 16$\times$ larger pixel size) to be collected within a much shorter period of time and then computationally refined to nanometer detail. To our knowledge, it is the first approach explicitly designed for large-factor EM super-resolution in real-world settings. ZoomSEM employs a recurrent transformer architecture incorporating an affine-deformable multi-scale alignment (ADMA) module. This design leverages the inherent inter-section correlations in volumetric EM data: the ADMA module aligns axial features to integrate information from adjacent EM sections, and the recurrent transformer aggregates these aligned features across depth to model long-range dependencies. This yields more faithful reconstructions of the current section, especially improving the continuity and fine detail of elongated neurites and synapses that may span many sections.

A key factor in ZoomSEM's superiority is the real-world dataset we constructed. The resulting resource, RealSEM-Pair, consists of precisely registered low- and high-resolution SEM image pairs acquired from brain tissue, and faithfully captures authentic noise profiles, structural variations, and imaging artifacts typical of practical SEM acquisitions. This real-world dataset overcomes the limitations of synthetic training data, which often fail to replicate the true degradation patterns encountered in practice, thereby enabling ZoomSEM to learn robust and generalizable representations. Beyond supporting this study, RealSEM-Pair provides a valuable resource for the broader EM community, facilitating rigorous benchmarking and accelerating the development, validation and comparison of new computational methods grounded in real imaging conditions.

Our experiments demonstrate that ZoomSEM faithfully recovers ultrastructural features from low-resolution scans. On large-volume brain tissue, the network accurately reconstructs delicate structures such as dendritic spines, synapses, synaptic vesicles, mitochondria and other subcellular organelles (verified against high-resolution ground truth), even for 16$\times$ super-resolution. The network also exhibits strong cross-sample and cross-species generalization, with minimal performance degradation in downstream segmentation tasks compared to direct high-resolution imaging. In a breast-cancer organoid system, microvilli and other subcellular structures are accurately restored with minimal network refinement. Importantly, this translates into drastic speedup of high-resolution data acquisition: for example, imaging a volume at 16$\times$16 coarser sampling and then applying ZoomSEM yields comparable quality to standard-resolution imaging, but with roughly 256$\times$ less acquisition time (neglecting overhead). In practical terms, ZoomSEM can reduce imaging duration by nearly 30-fold, shortening a multi-month imaging campaign to days. As a result, our method can significantly enhance the capability of conventional single-beam SEM systems, typically costing hundreds of thousand US dollars, enabling their imaging speeds comparable to state-of-the-art multibeam SEMs priced at several million dollars. This not only accelerates data acquisition but also greatly lowers financial barriers, democratizing high-throughput, large-volume electron microscopy for a far broader community across connectomics and the life sciences. By lifting the resolution bottleneck, our method enables high-throughput connectome mapping and other large-volume EM applications that were previously impractical.

\section*{Results}
\subsection*{Workflow of using ZoomSEM for SEM super-resolution}\label{subsec2_1}

\begin{figure*}[tp]
    \centering
    \includegraphics[width=1\linewidth]{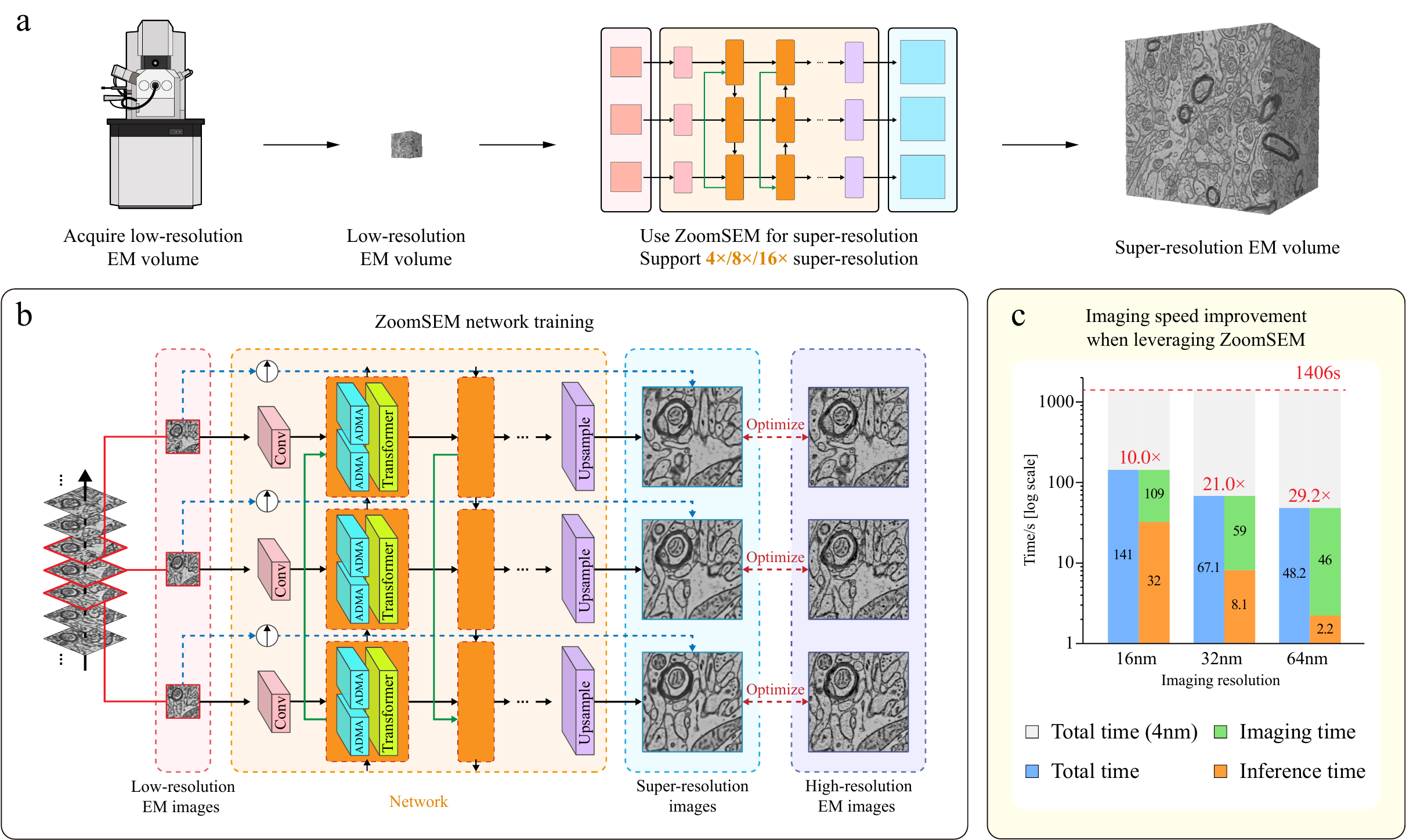}
    \caption{\textbf{Overview of the ZoomSEM network and speed advantages.}
\textbf{a}, Workflow of SEM super-resolution using ZoomSEM. Given a low-resolution SEM volume, the ZoomSEM network performs 4$\times$, 8$\times$, or 16$\times$ super-resolution without requiring high-resolution acquisition.
\textbf{b}, Training pipeline and schematic structure of the ZoomSEM network.
\textbf{c}, Estimated improvement in imaging speed when using ZoomSEM. Considering both acquisition and inference time, the overall speedup is 10.0-fold from 16\,nm to 4\,nm, 21.0-fold from 32\,nm to 4\,nm, and 29.2-fold from 64\,nm to 4\,nm. Time estimates are based on a 60\,\ensuremath{\mu}m $\times $60\,\ensuremath{\mu}m imaging area, corresponding to an image resolution of 15000$\times$15000 pixels at 4\,nm voxel size.
The inference time is measured using 8 A100 GPUs.}
\label{fig:fig0}
\end{figure*}

We aim to develop a method that directly predicts high-resolution EM images from low-resolution inputs using a trained network, thereby reducing both imaging cost and acquisition time (Fig.~\ref{fig:fig0}a).
This requires both representative training data and an architecture capable of exploiting the 3D spatial structure of serial EM volumes.
Existing datasets are limited to small upscaling factors or rely on synthetic degradation that poorly reflects real EM acquisition.
We therefore constructed RealSEM-Pair, a 3D serial SEM dataset from mouse entorhinal cortex containing pixel-level aligned volume pairs at four resolutions (4\,nm, 16\,nm, 32\,nm, and 64\,nm), registered via a stepwise strategy.
Even across a 16-fold resolution gap, our stepwise registration strategy achieves precise pixel-level alignment.
Compared to prior datasets, RealSEM-Pair provides up to 16-fold resolution spans from exclusively real acquisitions, with full axial registration and substantially larger scale.
Trained on this dataset, ZoomSEM employs a recurrent transformer architecture with bidirectional propagation across serial sections, and introduces the ADMA module to handle the structural scale variances characteristic of axial adjacent EM sections (Fig.~\ref{fig:fig0}b).

As the input resolution decreases, the acceleration gains increase markedly: 16\,nm imaging combined with ZoomSEM achieves a 10.0$\times$ speedup, 32\,nm imaging combined with ZoomSEM achieves a 21.0$\times$ speedup, and 64\,nm imaging combined with ZoomSEM achieves a 29.2$\times$ speedup (Fig.~\ref{fig:fig0}c).
These speed improvements account for both acquisition time and model inference time, measured using 8 A100 GPUs to ensure fair and reasonable computational resources.
Importantly, imaging speed remains the primary bottleneck in large-scale SEM workflows.
A substantial portion of the total imaging time is resolution-independent, including delays from stage movement, coil and field switching, scintillator and photodetector signal conversion, and data buffering or software processing.

\subsection*{ZoomSEM excels in SEM super-resolution across various upscaling factors}\label{subsec2_2}
\begin{figure*}[tp]
    \centering
    \includegraphics[width=1\linewidth]{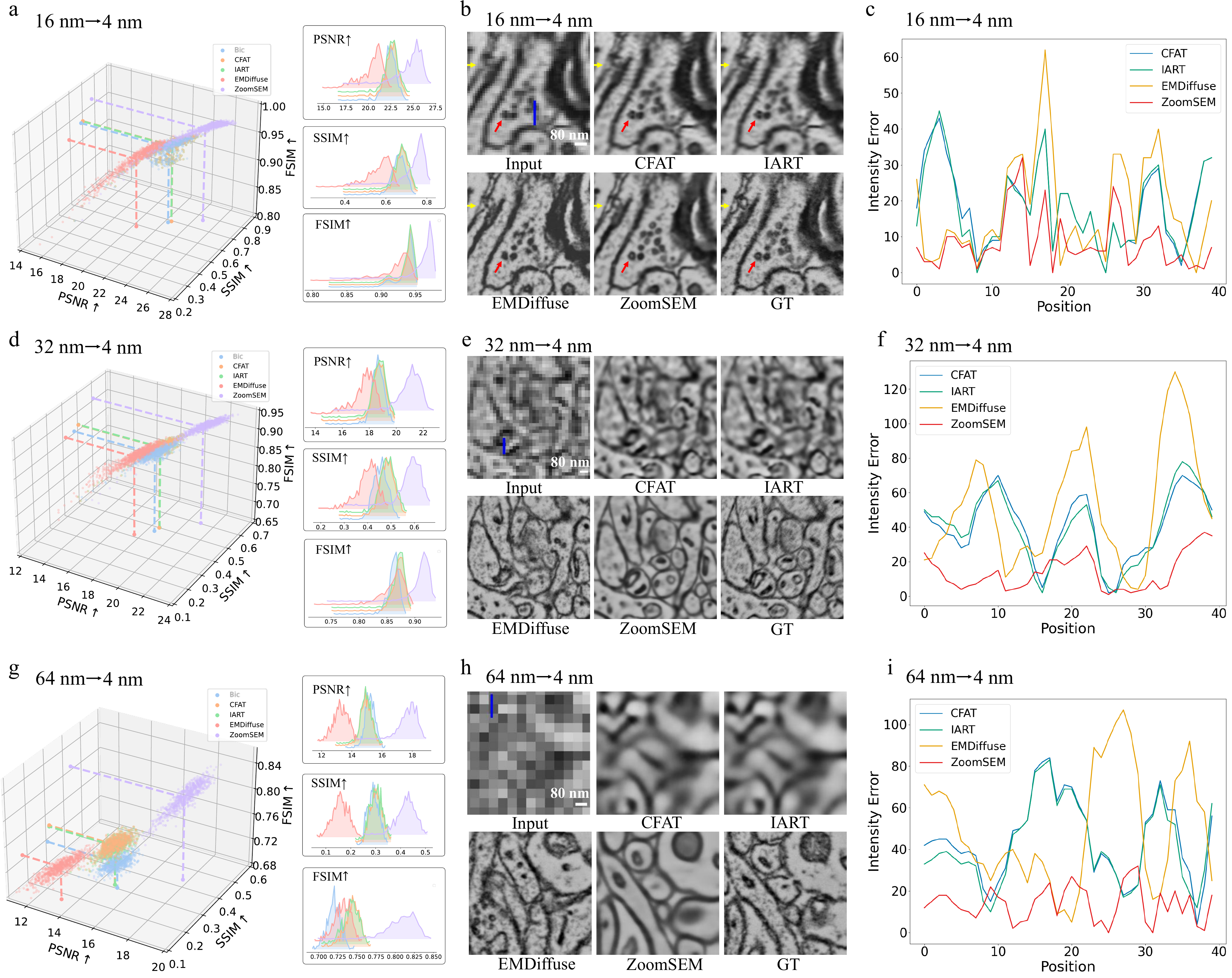}
  \caption{{\textbf{ZoomSEM achieves high-fidelity performance across various upscaling factors.}
\textbf{a,d,g}, Quantitative evaluation of ZoomSEM, CFAT \cite{ray2024cfat}, IART \cite{xu2024enhancing}, EMDiffuse \cite{lu2024diffusion}, and bicubic-interpolated inputs (Bic) for super-resolution tasks from 16\,nm to 4\,nm (4$\times$), from 32\,nm to 4\,nm (8$\times$), and from 64\,nm to 4\,nm (16$\times$), assessed by PSNR, SSIM \cite{wang2004image}, and FSIM \cite{zhang2011fsim}. All models are trained under their original training-data pipelines.
\textbf{b,e,h}, Qualitative comparison for super-resolution tasks from 16\,nm to 4\,nm (4$\times$), from 32\,nm to 4\,nm (8$\times$), and from 64\,nm to 4\,nm (16$\times$).
Arrows in b highlight fine structural details that are restored by our method but not by other methods.
From 16\,nm to 4\,nm, ZoomSEM closely matches the high-resolution ground truth (GT).
From 32\,nm to 4\,nm, ZoomSEM shows a pronounced advantage, successfully resolving fine structures that other methods fail to reconstruct while maintaining high fidelity.
From 64\,nm to 4\,nm, other methods largely fail to recover meaningful content. Despite minor errors in the internal cellular structures, ZoomSEM recovers the overall cell morphology, even when the input retains almost no discernible details.
\textbf{c,f,i}, pixel-wise intensity error comparison along the blue lines in panels b, e, and h. ZoomSEM is compared with baseline methods by measuring the pixel intensity differences along the blue lines, with GT as reference.}
\label{fig:fig1}
}
\end{figure*}

We comprehensively evaluate ZoomSEM on three super-resolution tasks: from 16\,nm to 4\,nm (4$\times$), from 32\,nm to 4\,nm (8$\times$), and from 64\,nm to 4\,nm (16$\times$). Three strong baselines are included for comparison: CFAT (single image super-resolution method) \cite{ray2024cfat}, IART (video super-resolution method) \cite{xu2024enhancing}, and EMDiffuse (EM super-resolution method) \cite{lu2024diffusion}.
We use PSNR, SSIM \cite{wang2004image}, and FSIM \cite{zhang2011fsim} for quantitative assessments. LPIPS \cite{zhang2018unreasonable} is excluded due to its emphasis on perceptual texture similarity over structural fidelity, which misaligns with the objective of accurately replicating true high-resolution SEM images of cellular structures.
The test set comprises four volumes, each containing 177 sections at 1536$\times$1536 pixels, totalling 708 high-resolution images.
As shown in Fig.~\ref{fig:fig1}a, d, and g, ZoomSEM consistently outperforms all baselines in all fidelity metrics.

We further assess qualitative outcomes by comparing input and output images (Fig.~\ref{fig:fig1}b, e, and h).
For super-resolution from 16\,nm to 4\,nm (4$\times$), ZoomSEM closely matches the high-resolution ground truth (GT) and is the only method that accurately reconstructs vesicle locations (Fig.~\ref{fig:fig1}b).
For super-resolution from 32\,nm to 4\,nm (8$\times$), ZoomSEM maintains high fidelity and resolves fine structures other methods fail to capture (Fig.~\ref{fig:fig1}e).
At the extreme 16$\times$ magnification (from 64\,nm to 4\,nm), ZoomSEM recovers most cellular features, despite minor errors, whereas other approaches largely fail (Fig.~\ref{fig:fig1}h).
Although EMDiffuse produces outputs stylistically closer to the GT, it introduces numerous incorrect structures.
In contrast, ZoomSEM accurately reconstructs cellular ultrastructure, rendering it more suitable for real-world, large-scale EM super-resolution tasks.
We also visualize the pixel-wise intensity error in Fig.~\ref{fig:fig1}c, f, i, showing that ZoomSEM consistently produces pixel values closer to the GT than existing methods across all three super-resolution settings.

\subsection*{Advancing SEM super-resolution with real-world training dataset and superior network architecture}\label{subsec2_3}
\begin{figure*}[tp]
    \centering
    \includegraphics[width=1\linewidth]{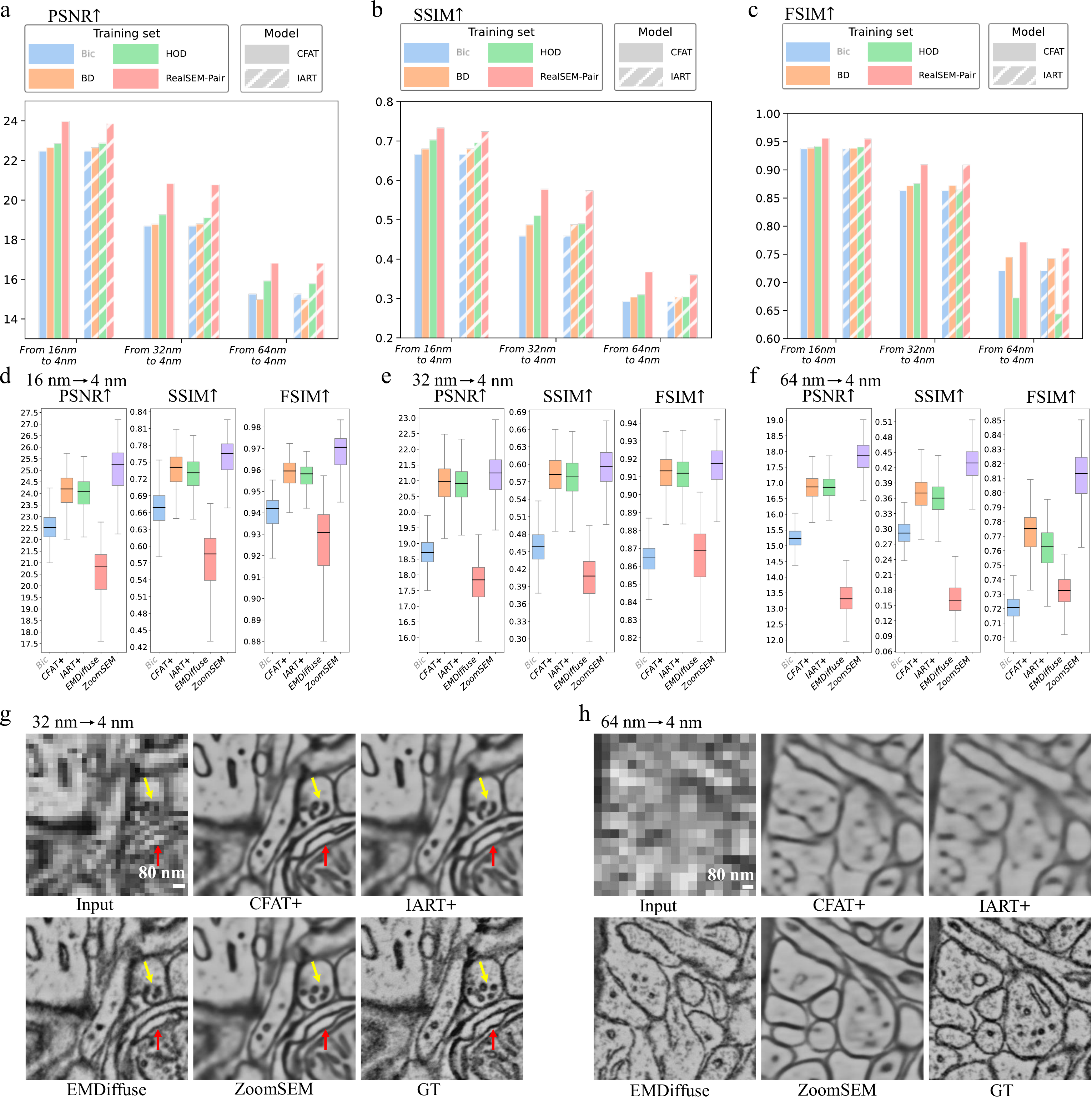}
  \caption{\textbf{Training on the RealSEM-Pair dataset and employing the ZoomSEM network substantially improve EM super-resolution performance.}
\textbf{a}, Schematic of the RealSEM-Pair dataset creation pipeline, including data preparation and processing.
\textbf{b}, Samples from the RealSEM-Pair dataset, showing pixel-level alignment of 3D SEM volumes at different resolutions, which demonstrates the feasibility of our registration strategy.
\textbf{c-e}, Quantitative comparisons from 16\,nm to 4\,nm (4$\times$), from 32\,nm to 4\,nm (8$\times$), and from 64\,nm to 4\,nm (16$\times$) super-resolution on the same models trained with the RealSEM-Pair dataset, the bicubic downsampling (BD) pipeline (synthetic), and the high-order downsampling pipeline (HOD) (synthetic) \cite{wang2021real}. Bicubic interpolation (Bic) is also provided for reference. The models used here are CFAT \cite{ray2024cfat} and IART \cite{xu2024enhancing}, evaluated by PSNR, SSIM \cite{wang2004image}, and FSIM \cite{zhang2011fsim}.
\textbf{f-h}, Quantitative comparisons from 16\,nm to 4\,nm (4$\times$), from 32\,nm to 4\,nm (8$\times$), and from 64\,nm to 4\,nm (16$\times$) super-resolution under training on the RealSEM-Pair dataset, evaluating ZoomSEM, CFAT+ \cite{ray2024cfat}, IART+ \cite{xu2024enhancing}
, EMDiffuse \cite{lu2024diffusion}, and Bic.
``+'' indicates that the model is trained on the RealSEM-Pair dataset, discarding its original synthetic data training pipeline.
Since EMDiffuse is already trained on the RealSEM-Pair dataset, EMDiffuse+ is identical to EMDiffuse.
\textbf{i}, Qualitative comparison from 32\,nm to 4\,nm (8$\times$) super-resolution. Other methods exhibit noticeable artifacts, whereas ZoomSEM demonstrates superior fidelity.
\textbf{j}, Qualitative comparison from 64\,nm to 4\,nm (16$\times$) super-resolution. At this extreme upscaling factor, while other methods largely fail, ZoomSEM recovers the overall cell morphology.
Arrows in panel i highlight fine structural details that are restored by our method but not by other methods.
}
\label{fig:fig2}
\end{figure*}

To validate the necessity of real-world training data, we compared models trained on synthetic degradation pipelines---bicubic downsampling (BD) and the high-order degradation pipeline (HOD) \cite{wang2021real}---against those trained directly on real image pairs.
Models trained on synthetic data perform notably worse under real super-resolution scenarios (Fig.~\ref{fig:fig2}a-c), confirming that real paired data are essential for effective EM super-resolution.

To demonstrate that the proposed ZoomSEM network itself offers inherent advantages, we trained all baseline methods on the RealSEM-Pair dataset using PSNR, SSIM \cite{wang2004image}, and FSIM \cite{zhang2011fsim} as quantitative metrics, and compared results with CFAT \cite{ray2024cfat}, IART \cite{xu2024enhancing}, and EMDiffuse \cite{lu2024diffusion}.
As illustrated in Fig.~\ref{fig:fig2}d-f, ZoomSEM markedly outperforms these baselines in terms of numerical fidelity.
We further perform qualitative comparisons of input and output images (Fig.~\ref{fig:fig2}g-h).
For the 32\,nm to 4\,nm (8$\times$) task, CFAT+ and IART+ (cf.\ Fig.~\ref{fig:fig1}e and Fig.~\ref{fig:fig2}g) visibly improve upon CFAT and IART, owing to training on our RealSEM-Pair dataset; nevertheless, ZoomSEM demonstrates superior structural fidelity.
In the more challenging 64\,nm to 4\,nm (16$\times$) task, although most methods fail to reconstruct the underlying architecture, ZoomSEM successfully recovers the global cellular morphology (Fig.~\ref{fig:fig2}h).
Notably, while IART also employs a recurrent transformer architecture, its alignment module does not fully leverage the spatial correlations in EM images.
By contrast, our ADMA module incorporates both affine transformations and sub-pixel-level alignment across EM sections, leading to results that significantly surpass those of IART+.
EMDiffuse is already trained on RealSEM-Pair; consequently, no additional dataset refinements are applied in this comparison.

\begin{figure*}[tp]
    \centering
    \includegraphics[width=1\linewidth]{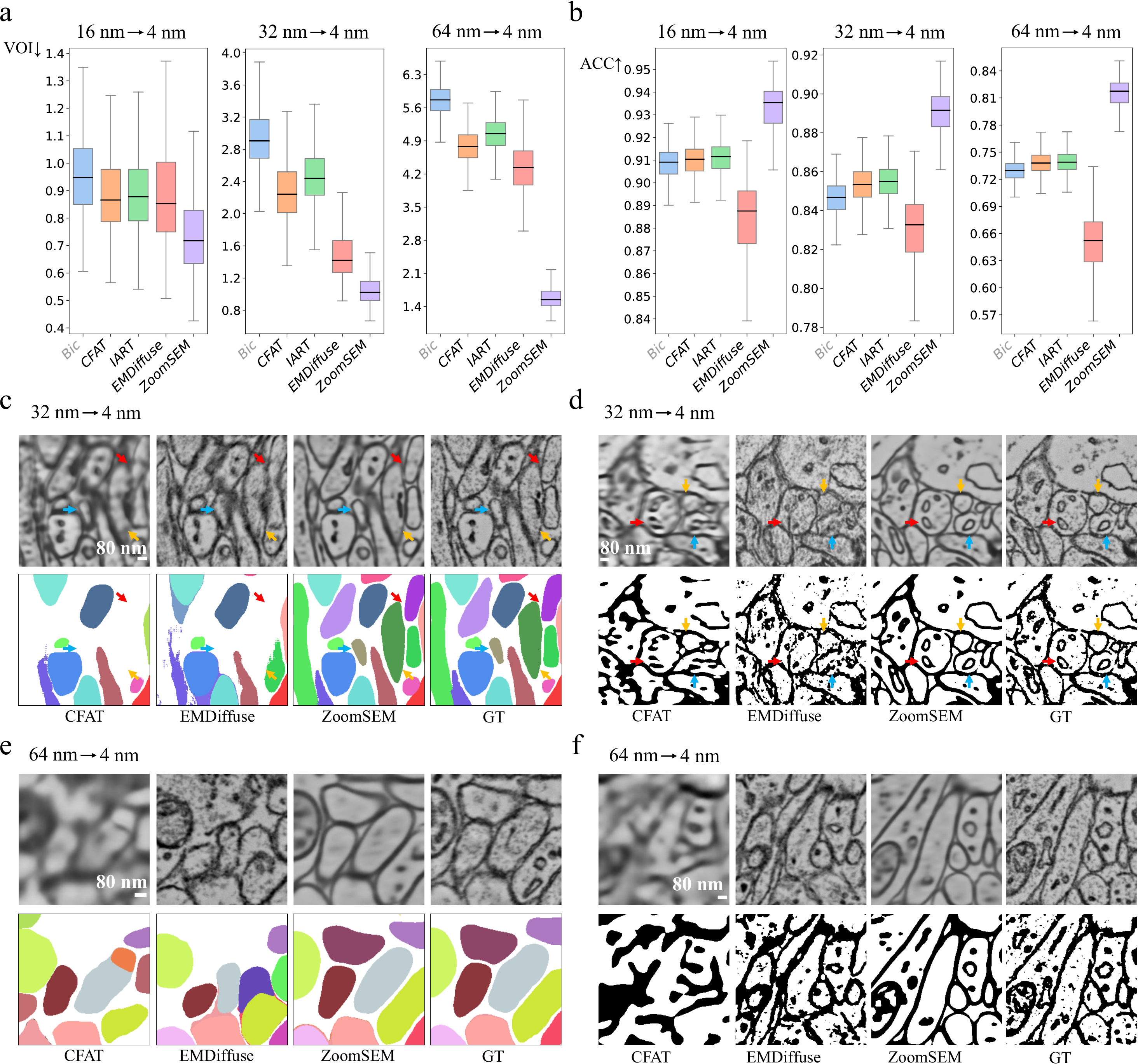}
  \caption{\textbf{Accurate downstream segmentation of ZoomSEM demonstrated via SAM-\cite{kirillov2023segment} and Otsu-based \cite{otsu1975threshold} segmentation.}
\textbf{a}, Quantitative evaluation of ZoomSEM, CFAT \cite{ray2024cfat}, IART \cite{xu2024enhancing}, EMDiffuse \cite{lu2024diffusion}, and bicubic-interpolated inputs (Bic) using SAM-based segmentation against SAM-generated pseudo-labels from the ground truth (GT). The results from 16\,nm to 4\,\,nm (4$\times$), from 32\,nm to 4\,nm (8$\times$), and from 64\,nm to 4\,nm (16$\times$) super-resolution tasks are assessed by VOI \cite{nunez2013machine}.
\textbf{b}, Quantitative evaluation of the same models using Otsu-based segmentation against Otsu-generated pseudo-labels from the GT, across the same three EM super-resolution settings, assessed by ACC.
\textbf{c-d}, Qualitative comparison of SAM- and Otsu-based segmentations from 32\,nm to 4\,nm (8$\times$) super-resolution task. ZoomSEM produces a clearer, more complete cellular structure with significantly higher semantic consistency than other methods.
Arrows in all panels (c-d) indicate fine structural details that are accurately super-resolved and segmented by our method but are either missing or misrepresented in the results of other methods.
In panel c, the SAM result is further refined by supplying an identical foreground prompt to every input patch, compensating for slight segmentation error in GT.
\textbf{e-f}, Qualitative comparison of SAM- and Otsu-based segmentations from 64\,nm to 4\,nm (16$\times$) super-resolution task. At this extreme scale, other methods fail to recover reliable structures, while ZoomSEM reconstructs coherent cellular morphology, yielding segmented results close to the GT. All segmentation was performed fully automatically without manual correction.
}
\label{fig:fig3}
\end{figure*}

\subsection*{Accurate downstream segmentation enabled by ZoomSEM}\label{subsec2_4}

To validate the accuracy of ZoomSEM reconstructions in downstream segmentation, we utilized both the Segment Anything Model (SAM) \cite{kirillov2023segment}, a large deep-learning-based segmentation model, and the conventional Otsu thresholding method \cite{otsu1975threshold}, to automatically segment the EM super-resolution outputs of all methods.
Specifically, we employed VOI \cite{nunez2013machine} to assess SAM-based segmentation results and defined the pixel-level accuracy (ACC) metric to measure consistency with a binarized GT for Otsu-based segmentation.
All segmentation procedures were fully automated with no manual refinements.
Since the segmentation labels were pseudo-labels generated automatically from the GT, a small degree of error and inconsistencies in mask selection might arise.
Nonetheless, the segmentation metrics remained reliable for evaluating overall consistency with the GT.
Two key conclusions emerge from these experiments (Fig.~\ref{fig:fig3}).
First, ZoomSEM demonstrates significantly higher semantic consistency with the GT across both SAM- and Otsu-based evaluations compared to other methods.
Second, this indicates that ZoomSEM's outputs more closely align with the GT at the image level.

\subsection*{Robust cross-sample and cross-species generalization of ZoomSEM}\label{subsec2_5}

\begin{figure*}[tp]
    \centering
    \includegraphics[width=0.92\linewidth]{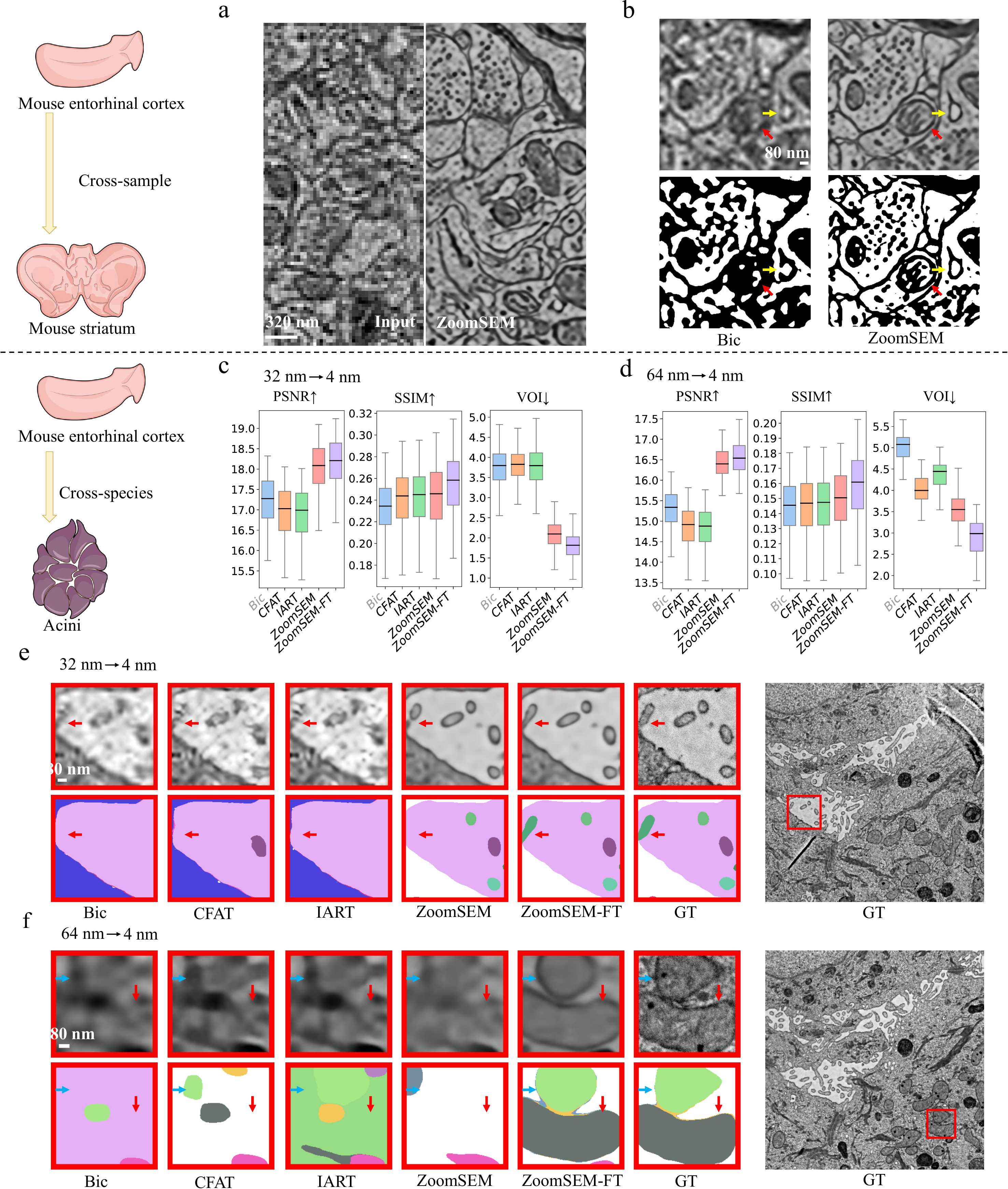}
  \caption{\textbf{ZoomSEM demonstrates excellent generalization performance.}
\textbf{a}, ZoomSEM super-resolves images from 32\,nm to 4\,nm (8$\times$). The network is trained on the RealSEM-Pair dataset (sampled from mouse entorhinal cortex) and evaluated on large-scale mouse striatum data. The comparison shows the low-resolution input and the high-resolution output.
\textbf{b}, Qualitative comparison of Otsu-based segmentation results between ZoomSEM and bicubic-interpolated inputs (Bic) on large-scale mouse striatum data from 32\,nm to 4\,nm (8$\times$). ZoomSEM can super-resolve clear cell boundaries and intracellular organelle structures without any fine-tuning.
\textbf{c-d}, Quantitative evaluation of ZoomSEM-FT, ZoomSEM, CFAT \cite{ray2024cfat}, IART \cite{xu2024enhancing}, and Bic on the acini dataset from 32\,nm to 4\,nm (8$\times$) and from 64\,nm to 4\,nm (16$\times$) super-resolution tasks, measured by PSNR, SSIM \cite{wang2004image}, and VOI \cite{nunez2013machine}. ZoomSEM-FT denotes the network fine-tuned on the acini dataset, while the other models are trained from scratch on the acini dataset.
\textbf{e}, Qualitative comparison from 32\,nm to 4\,nm (8$\times$) super-resolution task and SAM-based segmentation on the acini dataset. ZoomSEM-FT outperforms ZoomSEM in both fidelity and segmentation consistency relative to the ground truth (GT), while ZoomSEM notably surpasses CFAT and IART.
\textbf{f}, Qualitative comparison from 64\,nm to 4\,nm (16$\times$) super-resolution task on the acini dataset. Under this more challenging setting, all other methods, including ZoomSEM, fail to preserve the overall structure, whereas ZoomSEM-FT achieves a marked improvement in fidelity and segmentation consistency with respect to the GT.
Arrows in all panels (b, e-f) indicate fine structural details that are accurately super-resolved and segmented by our method but are either missing or misrepresented in the results of other methods.}
\label{fig:fig4}
\end{figure*}

We then investigate the cross-sample and cross-species generalization capabilities of ZoomSEM. To evaluate cross-sample generalization under large-scale dataset, we used a 32\,nm low-resolution SEM dataset acquired from the mouse striatum. We selected a central region of interest with a field of view of 5120$\times$5120 pixels at 32\,nm resolution and acquired 900 serial sections. Two sections with poor imaging quality were excluded. After axial registration, patch extraction, and brightness normalization, inference was performed on the remaining 898 slices using a ZoomSEM network trained on the RealSEM-Pair dataset (from mouse entorhinal cortex), without any fine-tuning. The resulting super-resolved volume totaled approximately 795GB.
As shown in Fig.~\ref{fig:fig4}a, ZoomSEM significantly enhances the resolution of the low-resolution input volume. It reveals clear cell boundaries and intracellular organelle structures that are not visible in the original input. Moreover, Otsu-based binarization highlights structural features that are absent in the low-resolution data but successfully reconstructed by ZoomSEM (Fig.~\ref{fig:fig4}b).

\begin{figure*}[tp]
    \centering
    \includegraphics[width=1\linewidth]{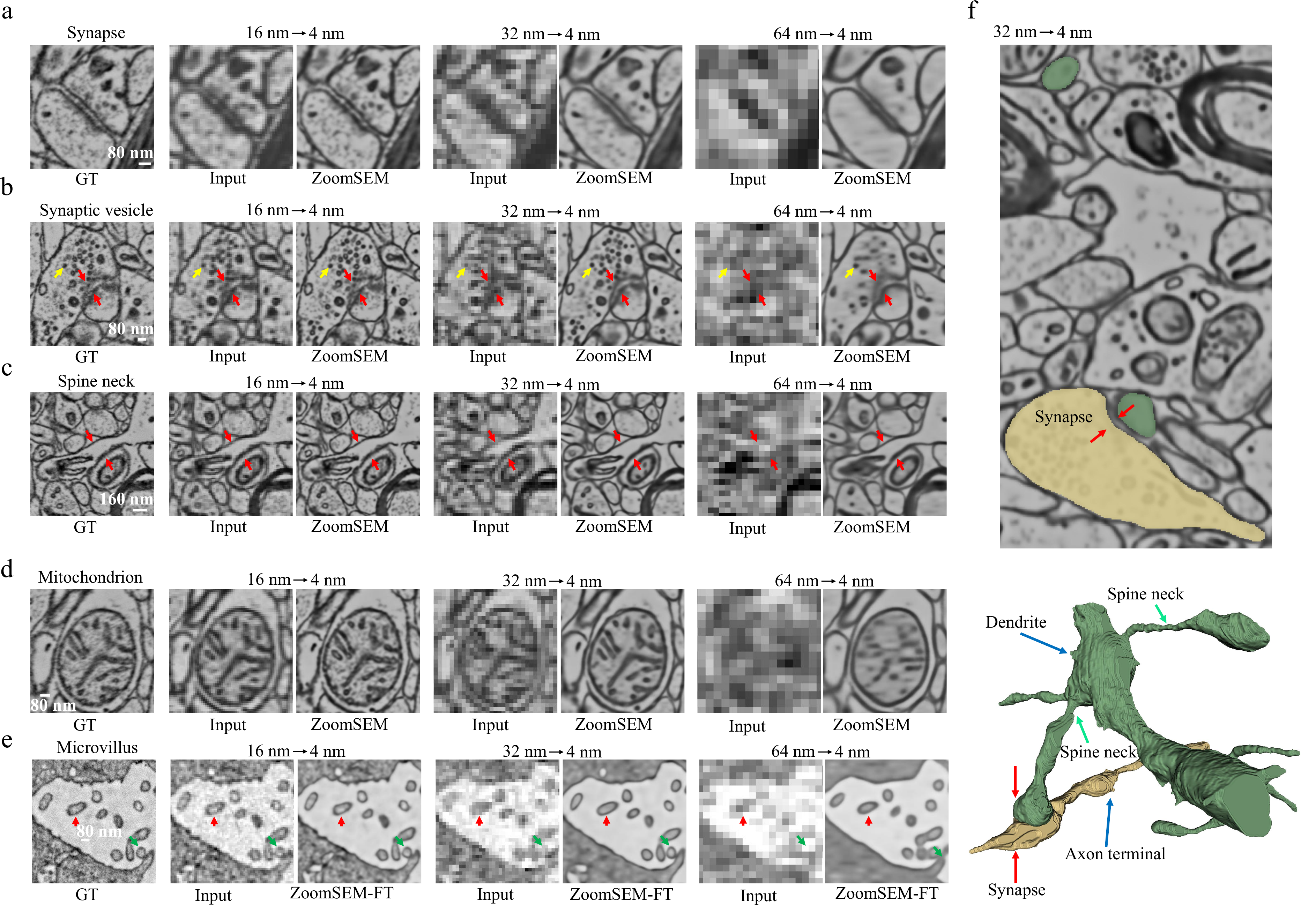}
  \caption{\textbf{ZoomSEM facilitates ultrastructure recovery.}
\textbf{a}, Synapses under three EM super-resolution settings: from 16\,nm to 4\,nm, from 32\,nm to 4\,nm, and from 64\,nm to 4\,nm. Comparing ZoomSEM to input, synaptic clefts transform from blurred or imperceptible to clearly defined and detailed from 16\,nm to 4\,nm, become distinguishable again from 32\,nm to 4\,nm (after being virtually lost), and emerge from near-complete absence at from 64\,nm to 4\,nm to reveal a recognizable synaptic outline.
\textbf{b}, Synaptic vesicles at the same three EM super-resolution scales. ZoomSEM faithfully reconstructs vesicles (yellow arrows) when super-resolving from 16\,nm and 32\,nm to 4\,nm. Although vesicles are unresolved in the 64\,nm to 4\,nm setting, the synaptic cleft is still clearly visible (between red arrows), so the synapse remains easy to recognize.
\textbf{c}, Spine neck at three EM super-resolution scales.  
The ultra-thin neck of the dendritic spine is successfully recovered in all three super-resolution settings.
\textbf{d}, Mitochondria at three EM super-resolution scales. ZoomSEM faithfully recovers mitochondrial cristae from 32\,nm to 4\,nm super-resolution.
\textbf{e}, Microvilli at three EM super-resolution scales using ZoomSEM-FT. Larger microvilli indicated by red arrows are accurately reconstructed across all scales, while finer microvillar structures (green arrows) are faithfully recovered from 16\,nm to 4\,nm and from 32\,nm to 4\,nm, but remain unresolved in the 64\,nm to 4 \,nm setting.
\textbf{f}, Manual tracing on the 32\,nm to 4\,nm super-resolved volume reconstructs an axon bouton-dendritic spine pair, and short segments of the axon terminal and dendrite are further traced. Left, a representative section with hand-drawn annotations. Right, the resulting 3D tracing.
}
\label{fig:fig5}
\end{figure*}

We also validated the cross-species generalization capability of ZoomSEM, expanding its utility from connectome mapping to broader life science applications. We constructed a small-scale human breast cell acini dataset.
The acini samples were derived from Jorgens \textit{et al.} \cite{jorgens2017deep}, and we re-imaged them under the same pipeline described above.
This dataset comprises 200 serial EM sections with pixel-level alignment across multiple resolutions.
We used the first 100 sections to fine-tune ZoomSEM, reserving the remaining 100 for testing; the fine-tuned network is referred to as ZoomSEM-FT.
We then compared ZoomSEM-FT with ZoomSEM, CFAT, and IART, each trained from scratch on 100 acini images, by evaluating their performance on the validation set from two perspectives: visual quality and consistency with the GT segmentation.
On the acini dataset, ZoomSEM already outperforms the two baseline methods in quantitative metrics.
Moreover, ZoomSEM-FT significantly surpasses ZoomSEM, especially in SSIM and VOI (Fig.~\ref{fig:fig4}c-d), suggesting that the fine-tuned network achieves substantially better structural similarity.
Qualitative comparisons further corroborate these findings (Fig.~\ref{fig:fig4}e-f).
Notably, at 64\,nm to 4\,nm (16$\times$) upscaling, other methods (including ZoomSEM) fail to reconstruct the more homogeneous texture of acini, whereas ZoomSEM-FT---benefiting from the data priors obtained from the RealSEM-Pair dataset---achieves marked improvements in fidelity and segmentation consistency with the GT.

These results highlight two key outcomes. First, ZoomSEM consistently outperforms all baselines, confirming that its performance advantages on the mouse entorhinal cortex data generalize to the acini dataset. Second, the fine-tuned variant ZoomSEM-FT further improves reconstruction quality, demonstrating the value of RealSEM-Pair as an effective training prior. In effect, ZoomSEM trained on the RealSEM-Pair dataset serves as a transferable foundation model, capable of generalizing to new imaging contexts. Notably, it achieves strong performance in cross-sample settings without any additional training data, and can be adapted to cross-species datasets with only a modest amount of registered image pairs (just 100 in this case).

\subsection*{Ultrastructure recovery with ZoomSEM}\label{subsec2_6}

We assess the practical implications of ZoomSEM by analyzing its output image quality on key subcellular structures (including synapses, synaptic vesicles, spine necks, mitochondria and microvilli), its impact on downstream 3D neuron reconstruction accuracy, and the acceleration of imaging throughput achieved by leveraging ZoomSEM.

Applying super-resolution techniques to visualize synaptic structures with greater clarity is crucial for uncovering the molecular mechanisms underlying neuronal communication, synaptic plasticity, and pathology \cite{sudhof2013neurotransmitter, nicoll2017brief}.
Across 16\,nm to 4\,nm, 32\,nm to 4\,nm, and 64\,nm to 4\,nm super-resolution tasks, ZoomSEM enhances synaptic clefts from being blurred or imperceptible to clearly distinguishable structures, with fine details at 16\,nm, recoverable synapses at 32\,nm, and even recognizable cellular contours at 64\,nm (Fig.~\ref{fig:fig5}a).

Applying super-resolution to accurately visualize synaptic vesicles is paramount for elucidating the molecular mechanisms of neurotransmitter release, vesicle recycling, and activity-dependent membrane trafficking at synapses \cite{rothman1994mechanisms, rizzoli2005synaptic}.
With ZoomSEM super-resolution, synaptic vesicles can be visualized distinctly and correctly (Fig.~\ref{fig:fig5}b), enabling direct inspection of vesicle pools within a single axon bouton.

Dendritic spines---key postsynaptic sites for excitatory signaling---also benefit markedly from ZoomSEM.
Resolving their morphology is essential for understanding synaptic plasticity and biochemical compartmentalization \cite{rochefort2012dendritic}.
Notably, the ultra-thin spine neck, which is usually the most difficult substructure to discern, is faithfully recovered under all three super-resolution settings (Fig.~\ref{fig:fig5}c).

Mitochondria are another key structure; their cristae organize the oxidative-phosphorylation machinery to sustain adenosine triphosphate production \cite{cogliati2013mitochondrial,cogliati2016mitochondrial}, and ZoomSEM faithfully recovers cristae from 32\,nm to 4\,nm super-resolution (Fig.~\ref{fig:fig5}d).

Applying super-resolution to precisely visualize microvillar architecture is essential for elucidating their roles in epithelial absorption and mucosal defence, as microvilli are actin-supported apical-membrane protrusions that assemble into an ordered brush border, thereby expanding surface area and enriching transporters and enzymes \cite{sauvanet2015structure,miura2015fluid,shifrin2012enterocyte}.
ZoomSEM-FT restores microvillar structures that are indiscernible at low-resolution imaging (Fig.~\ref{fig:fig5}e). Specifically, densely packed microvillar regions remain unresolvable at 32\,nm imaging but are successfully reconstructed by ZoomSEM-FT (green arrow, Fig.~\ref{fig:fig5}e).
Larger microvillar structures ($\sim$200\,nm diameter) are indistinguishable at 64\,nm imaging but are accurately reconstructed by ZoomSEM-FT (red arrow, Fig.~\ref{fig:fig5}e).

Finally, we selected the 32\,nm to 4\,nm super-resolution model as a compromise between imaging throughput and reconstruction fidelity.
Using this model, we manually traced and reconstructed an axon bouton-dendritic spine pair, including short segments of the axon terminal and dendrite. As shown in Fig.~\ref{fig:fig5}f, even the thinnest spine neck is faithfully recovered, demonstrating the feasibility of our approach for practical imaging applications.

In summary, ZoomSEM significantly accelerates imaging while maintaining high reliability, demonstrating its potential to transform EM imaging workflows, enhance biological research efficiency, and drive scientific discovery.

\section*{Discussion}

Serial electron microscopy serves as the cornerstone for mapping large-scale neural circuits at nanometer resolution but is hindered by severe throughput limitations. Standard single-beam scanning electron microscopes require months or even years to image brain volumes at scales of hundreds of micrometers. Advanced multi-beam SEM systems can image substantially faster but remain largely inaccessible due to high financial barriers. This limitation significantly constrains large-scale connectomic studies, making high-resolution serial EM challenging to use for many research groups.

In this paper, we introduce ZoomSEM, a deep learning-based method that restores high-resolution SEM images from rapidly acquired low-resolution scans, enabling substantial acceleration without sacrificing structural fidelity. Unlike conventional super-resolution methods trained on synthetic datasets, ZoomSEM leverages a unique, real-world dataset, RealSEM-Pair, comprising precisely registered low- and high-resolution EM image pairs. The use of real-world paired data is critical, as low-resolution EM images capture authentic degradation characteristics, including noise, aberrations, and staining variations, which cannot be faithfully replicated by synthetic data. By combining real-world training data with an advanced transformer-based architecture, ZoomSEM effectively reconstructs ultrastructural features from initial resolutions up to $16\times$ coarser, translating into data acquisition speedup of approximately 10 to nearly 30 times. This dramatically improves the utility of standard single-beam SEMs, enabling their imaging performance to approach costly multi-beam instruments and democratizing access to high-throughput volumetric EM.
To our knowledge, this is the first approach tailored for large-factor EM super-resolution in real-world settings.

ZoomSEM reliably preserves fine biological details, including synapses, synaptic vesicles, spine necks, mitochondria, microvilli, and organelle morphologies across different species and tissue types, demonstrating robust cross-domain generalization, underscoring its potential as a broadly applicable tool in neurobiology and cell biology. Our RealSEM-Pair dataset itself represents a valuable resource for the EM community, enabling rigorous benchmarking and training of future computational methods on authentic imaging data.

The accelerated imaging provided by ZoomSEM substantially impacts downstream analyses, notably automated segmentation and connectome reconstruction. By producing high-fidelity reconstructions from low-resolution input, ZoomSEM consistently achieves higher consistency with ground truth (GT) segmentation than baseline methods, particularly at larger upscaling factors. Even in challenging cases where other approaches fail to recover recognizable cellular structures, ZoomSEM preserves key ultrastructural features with sufficient accuracy to support reliable instance interpretation. This advancement enables researchers to undertake comprehensive brain-wide mapping or investigate ultrastructural pathology at scales previously impractical with existing SEM hardware.

Despite its strengths, ZoomSEM has boundaries that merit consideration. The method is currently demonstrated on SEM data. Extending it to other EM modalities might require additional training data and potential architectural adjustments. Image artifacts, such as section defects, charging, or contamination, etc., are not explicitly handled. Moreover, although a model trained on mouse entorhinal cortex generalizes reasonably well to mouse striatum, its performance remains inferior to a counterpart fine-tuned directly on striatal data---reflecting domain gaps introduced by sample preparation, imaging hardware, and brain-region-specific contrast. To deploy ZoomSEM on other specimens or microscopes, one can follow our training pipeline to acquire a modest number of paired images and fine-tune the released model, thereby improving accuracy on the target domain. Generalizing to extremely low signal conditions (e.g. ultra-fast scans with very low dose) may also demand further model robustness. Future efforts will extend ZoomSEM's capabilities to more diverse EM modalities, imaging conditions and sample types. Incorporating artifact detection or robust loss functions could further improve reliability. Additionally, exploring unsupervised or semi-supervised learning approaches may help reduce dependence on paired ground truth data. Finally, integrating ZoomSEM with downstream segmentation tools into fully automated pipelines could ultimately establish end-to-end workflows for rapid, accurate and scalable connectomic and cellular imaging studies.

In summary, ZoomSEM significantly redefines the speed and scale at which volumetric EM imaging can be conducted. By bridging the gap between high-resolution quality and rapid imaging, ZoomSEM stands to transform the accessibility and feasibility of ambitious connectomics and ultrastructural biology studies.

\bibliographystyle{unsrtnat}
\bibliography{references}
\end{document}